\documentclass{article}

\usepackage{todonotes}
\usepackage{xspace}
\usepackage{amsmath}
\usepackage{caption,subcaption}
\usepackage[]{algorithm2e}

\PassOptionsToPackage{numbers, compress}{natbib}
\usepackage[final]{nips_2018}

\usepackage[utf8]{inputenc} % allow utf-8 input
\usepackage[T1]{fontenc}    % use 8-bit T1 fonts
\usepackage{hyperref}       % hyperlinks
\usepackage{url}            % simple URL typesetting
\usepackage{booktabs}       % professional-quality tables
\usepackage{amsfonts}       % blackboard math symbols
\usepackage{nicefrac}       % compact symbols for 1/2, etc.
\usepackage{microtype}      % microtypography
\usepackage{float}
\usepackage{graphicx}

\title{A Framework for Implementing Machine Learning on Omics Data}

\author{
  Geoffroy Dubourg-Felonneau$^1$, Timothy Cannings$^1$$^,$$^2$, Fergal Cotter$^1$$^,$$^3$, \\
  \textbf{Hannah Thompson$^1$, Nirmesh Patel$^1$, John W Cassidy$^1$$^,$$^3$, Harry W Clifford$^1$} \\
  \\
  \and
  $^1$Cambridge Cancer Genomics \\
  $^2$University of Edinburgh \\
  $^3$University of Cambridge
}

\begin{document}

\maketitle

\begin{abstract}
The potential benefits of applying machine learning methods to -omics data are becoming increasingly apparent, especially in clinical settings. However, the unique characteristics of these data are not always well suited to machine learning techniques. These data are often generated across different technologies in different labs, and frequently with high dimensionality. In this paper we present a framework for combining -omics data sets, and for handling high dimensional data, making -omics research more accessible to machine learning applications. We demonstrate the success of this framework through integration and analysis of multi-analyte data for a set of 3,533 breast cancers. We then use this data-set to predict breast cancer patient survival for individuals at risk of an impending event, with higher accuracy and lower variance than methods trained on individual data-sets. We hope that our pipelines for data-set generation and transformation will open up -omics data to machine learning researchers. We have made these freely available for noncommercial use at \url{www.ccg.ai}.
\end{abstract}

\section{Introduction}

Cancer research has been revolutionized by the advent of high-throughput sequencing and the ability to generate data at an "-omics" level (genomics, transcriptomics, epigenomics, proteomics, etc). Implementation of machine learning techniques to -omics data is complex, however, when applied successfully they have been useful in obtaining meaningful biological insights. For example, although cancer is a highly heterogeneous disease with a diverse range of subtypes and clonal compositions \cite{Almendro2013CellularHA}, unsupervised learning has been successfully utilized to classify and interpret the unique genomic signatures found from one tumor to the next \cite{Dawson2013ANG}. This in turn has enabled stratification of patients into subgroups with distinct clinical outcomes. Another example is in prediction of drug-target interactions, for which machine learning is being used to narrow the search space for candidate drugs by application of predictive methods \cite{PMID30364884}.

These examples clearly demonstrate how useful machine learning can be in -omics, but have relied on the generation of -omics data specific to these purposes, rather than utilizing the vast amount of data that has already been generated. More commonly, this data is disparate, split across labs into small data sets, and generated with different technologies (e.g. RNA-Seq, microarrays). Additionally, the richness of -omics data enables extraction of a large number of features, which often outstrips the availability of patients and results in high dimensional data. Only large research labs capable of producing population-scale consistent -omics data can overcome these problems.

In this paper, we provide a pipeline for combining -omics data sets and methods for handling high dimensional data, making -omics research more accessible to machine learning applications (\autoref{sec:data}). To demonstrate the success of this, we use the combined data to predict breast cancer patient survival for individuals at risk of an impending event, with higher accuracy and lower variance than methods trained on individual data-sets (\autoref{sec:methods}). With the hope of this work enabling greater opportunity for implementing machine learning algorithms on -omics data produced in clinical and research settings, we have made these pipelines freely available for noncommercial use at \url{www.ccg.ai}.

\section{Data}\label{sec:data}

\subsection{Combining sources of RNA: gene expression and CNA: copy number aberration}

RNA is a proxy for gene expression. RNA data is usually found in the form of a $N \times M$ matrix, where $N$ is the number of patients and $M$ is the number of observed genes. Each value is a positive real number representing the level of expression for a given gene of a given patient, but these values differ considerably across technologies, each with their own biases and signal-noise distributions. Microarrays give an intensity value from RNA binding to probes, which roughly follows a gamma distribution (see Figure 1.1); RNA-Seq gives a count value from sequenced fragments of RNA, which follows a negative binomial distribution. We have been able to successfully combine 3 different datasets: METABRIC microarray, TCGA microarray, TCGA RNA-seq. The union of patients gives $N=3533$, and the intersection of genes gives $M=15233$, whilst retaining key characteristics, such as distinct disease-free survival in Integrative Cluster classification \cite{Dawson2013ANG}. 

\begin{figure}[H]
\centering
\includegraphics[width=6.6cm]{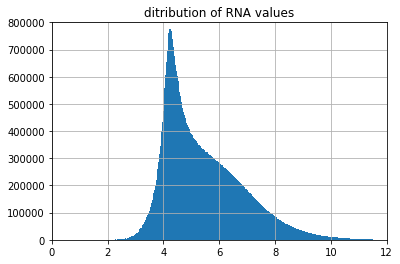}
\includegraphics[width=6cm]{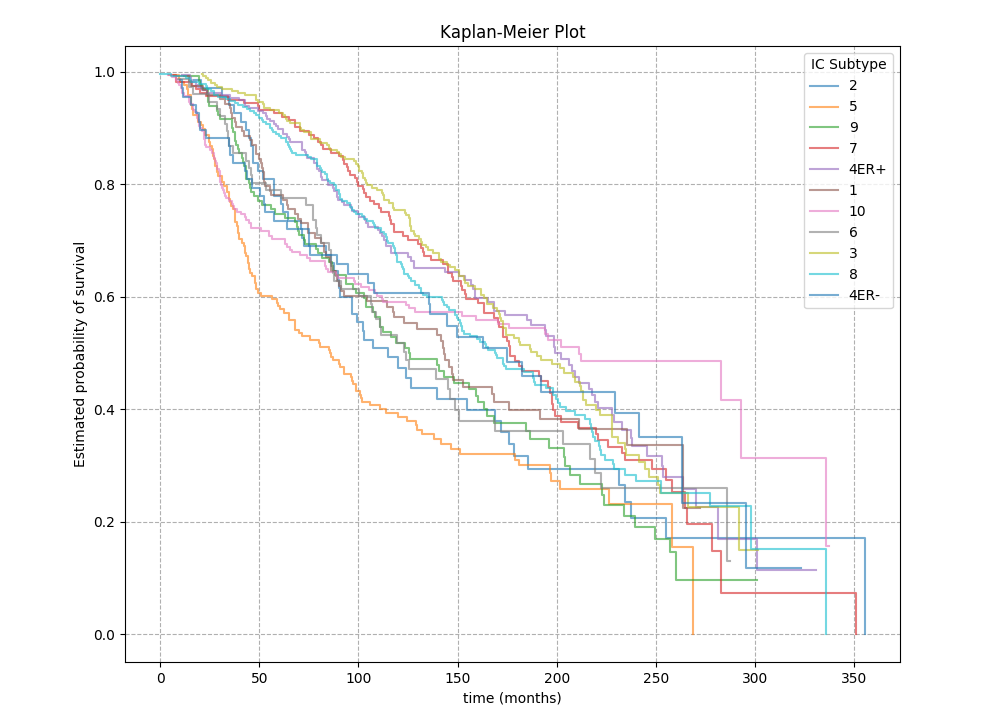}
\caption{(1) Combined RNA distribution. (2) Survival plots show Integrative Subtype retention.}
\end{figure}

The CNA data is found in the same dimensionality, with each value corresponding to the number of copies for a given region in the genome of a given patient. Both datasets were generated with GISTIC 2.0 \cite{Mermel2011} in the following categorical way: -2 = homozygous deletion; -1 = hemizygous deletion; 0 = neutral / no change; 1 = gain; 2 = high level amplification.

\subsection{Connecting clinical data and defining target variables}
\label{sec:target_vars}
We added patient age to the gene expression/copy number data to make our input $\mathbf{X}_i$. We defined the lifetime of patient $i$, as a random variable $T_i$ and wish to estimate the conditional probability of the \emph{survival function}:
$$S_T(t) = P(T > t | \mathbf{X})$$
where $t$ is time. Figure 1.2 shows survival functions for some patient groups in the dataset (grouping done by the unsupervised technique in \cite{Dawson2013ANG}). Due to the nature of clinical trials, not every patient is tracked until death. Many leave the study and their status becomes unknown. When this happens, we say the patient is \emph{lost} and the time they were last seen is $L_i$. If a patient remains in the study until death, then we have the true variable $T_i$. In our combined dataset, 55.4\% of the patients were tracked until death, with the remaining 44.6\% of patients lost. We define the clinical data for each patient as $C_i$, with
$$C_i = \min (T_i, L_i)$$

A common task is to estimate the probability at certain times, e.g. $S_T(60)$ or $S_T(24)$ (for 2 and 5 year survival). We can do this by building an estimator for $S_T$ directly and evaluating it at given times, or by converting the clinical data $C_i$ to classification labels $y_i$ and build a classifier. This simplifies the problem at the cost of coarseness in our prediction. We built $y_i$ with the following logic:
\begin{algorithm}[H]
 \KwData{Set t = 60 (5 years) or t = 24 (2 years)}
 \uIf{$C_i > t$}{
    $y_i = 1$ \tcp{patient survived at least $t$ months}
 }\uElseIf{$T_i <= L_i$ }{
    $y_i = 0$ \tcp{patient died before $t$}
 }\Else{
    drop \tcp{patient lost before $t$}
 }
\end{algorithm}

For both 2 and 5 year survival, there was significant class imbalance with 92\% ($ p(\mathbf{y}_i) = [0.08, 0.92]^T$) and 82\% ($p(\mathbf{y}_i) = [0.18, 0.82]^T$) of patients surviving respectively. Accuracy results are misleadingly high due to skewed prior distributions, so, we used receiver operating characteristic (ROC) curves to quantify our networks by calculating the area under the ROC (AUC). Values closer to 1 indicate a good classifier, and a random classifier will achieve an AUC of 0.5.

\subsection{Related work}\textbf{}
It is difficult to benchmark ourselves against other papers as many use much smaller sample sizes, different prediction targets, and few state the class split prior. However we do note that in many recent papers, an AUC for predicting 5 year survival of 0.75 is a good target \cite{mucaki_predicting_2017, urda_deep_2017}.
\section{Methods}\label{sec:methods}

To demonstrate increased performance in our combined data compared to baseline methods, we applied both supervised (we used our optimization pipeline to train supervised models on the raw labelled data) and semi-supervised techniques (we applied manifold learning techniques on all the data (with and without labels) to learn a way to project the data in a smaller representation space) on the same data for comparison. Then we applied classification methods on the labeled data projected in this space.

\subsection{Combining multiple data sources}

For combining disparate expression data-sets, we applied Feature Specific Quantile Normalization (FSQN) proposed for biological data by Franks JM et al \cite{Franks2018FeatureSQ}. We observed a significant increase in performance while training on the combined data rather than each dataset independently. For example, an MLP Classifier reaches 0.522 AUC when trained on METABRIC alone, but 0.754 AUC when trained on the combined dataset. Similarily, an SVC with RBF kernel reaches 0.657 AUC on METABRIC alone, but 0.81 on the combined dataset.

\subsection{Projections}

To address the issue of high dimensionality, we applied the t-sne \cite{vanDerMaaten2008} technique to project the  raw data into smaller spaces. Once again, we observed an increase in performance. For example, a Gaussian Process Classifier reaches .5 AUC on raw data, but 0.75 AUC on a 3D TSNE projection. For this reason, we integrated a projection module into the pipeline for iterations at multiple dimensions.

\subsection{Classifiers and Regressors}
For classification tasks, we use several standard classifiers such as Support Vector Machines, Naive Bayes, Lasso Regression and Random Forests. These could be applied to either the raw data (all $M$ genes), or to a projection of these genes. The Support Vector Classification with the Radial basis function kernel gave the best performance on the projected data (see \autoref{table:crossfoldsnp}). 

We applied a neural network regressor to build an estimator for $S_T(t)$. As mentioned in \ref{sec:target_vars}, many of the patients were lost from the study before death, so the variable $C_i$ is a lower bound on the true variable. We can account for this in the cost function by weighting down or even removing these samples, although we found it made little difference in the prediction accuracy. Instead, we simply minimize the mean squared error between the neural network output and $C_i$.
$$L = \frac{1}{N}\sum_{i=0}^{N-1} (C_i - f(x_i))^2$$
After fitting the network, we evaluated $S_T(60)$, and $S_T(24)$ and calculated the AUC (see \autoref{table:crossfoldsnp}). This achieved a lower overall AUC than the best classifiers but can be used to give us more complete data (e.g. we can generate a kaplan meier plot per patient with $S_T$).

Finally, we also applied the \textit{Random-projection ensemble classifier} proposed by \citet{Cannings:2017}. This method can be seen as a way to extend simple classifiers to high-dimensional data. Moreover, it allowed us to assess the relative importance of the features used in the prediction.  The classifier gave an AUC on 250 held-out test observations of 0.76 and 0.79 for 2 year and 5 year survival prediction, respectively. Tuning parameters where chosen using 10-fold cross validation.

\subsection{Pipeline}

We present a pipeline tool for easier experimentation and reproducibility in other data-sets across diseases. This pipeline allows us to perform the following; model wrapping - handling custom models that respect a simple interface; cross validation - automatic cross validation for evaluation; hyperparameter optimization - scanning a wide parameter space across multiple computational platforms using hyperopt \cite{hyperopt}; distribute the data - seamless data distribution across compute clusters. 

\section{Experiments and Results}
\begin{table}[h]
\caption{Results across different classifiers and data-sets}\label{table:crossfoldsnp}
\centering
\begin{tabular}{lllccccc}
\toprule
Model & Data & Validation AUC \\
\midrule
SVC (RBF) & TCGA Metabric RNA raw & 0.815\\
SVC (RBF) & TCGA Metabric RNA TSNE 15 age & 0.774\\ 
GaussianProcessClassifier & TCGA Metabric RNA TSNE 3 age & 0.755\\ 
RectangleMLPClassifier & TCGA Metabric RNA TSNE 40 age & 0.754\\
Lasso & TCGA Metabric RNA raw & 0.750 \\
GaussianNB & TCGA Metabric RNA TSNE 70 age & 0.742\\
SVC (RBF) & TCGA Metabric RNA TSNE 5 age & 0.736\\
GaussianProcessClassifier & TCGA Metabric RNA TSNE 5 age & 0.725\\
Neural Network Regressor & TCGA Metabric RNA raw & 0.720\\
GaussianNB & TCGA Metabric RNA TSNE 10 age & 0.692\\
Random Forest & TCGA Metabric RNA raw & 0.670 \\
SVC (RBF) & Metabric RNA raw & 0.662\\
GaussianNB & Metabric RNA raw & 0.657\\
GaussianProcessClassifier & TCGA Metabric RNA TSNE 70 age & 0.654\\ 
GaussianNB & TCGA Metabric RNA+CNA raw & 0.649\\ 
GaussianNB & TCGA Metabric RNA raw & 0.645\\ 
GaussianNB & TCGA Metabric CNA raw & 0.639\\ 
RectangleMLPClassifier & TCGA Metabric RNA raw & 0.607\\ 
RectangleMLPClassifier & TCGA Metabric RNA+CNA raw age & 0.551\\ 
RectangleMLPClassifier & Metabric RNA raw & 0.522\\ 
GaussianProcessClassifier & TCGA Metabric RNA raw & 0.500\\ 
\bottomrule
\end{tabular}
\end{table}
 
\section{Conclusion}

Our deep learning pipeline enables the use of high dimensional -omics data from disparate sources to predict clinical outcomes. We demonstrate this through prediction of short term survival in breast cancer patients, with the hope of greater monitoring and care for those patients at high risk. We believe this will be especially beneficial in opening up -omics data to machine learning researchers.

\bibliography{bib}

\end{document}